\documentclass[11pt,a4paper]{article}
\usepackage[hyperref]{naaclhlt2019}
\usepackage{times}
\usepackage{latexsym}
\usepackage[normalem]{ulem}
\useunder{\uline}{\ul}{}
\usepackage{url}

\aclfinalcopy 
\def\aclpaperid{1462} 

\newcommand{\vocab}{\mathcal{V}}
\newcommand{\vv}{\mathbf{v}}
\newcommand{\uu}{\mathbf{u}}

\newcommand{\ws}{\textsf{\small ws}}

\newcommand\blfootnote[1]{%
  \begingroup
  \renewcommand\thefootnote{}\footnote{#1}%
  \addtocounter{footnote}{-1}%
  \endgroup
}

\title{Better Word Embeddings by Disentangling Contextual n-Gram Information}

\author{ Prakhar Gupta*\\
  EPFL, Switzerland \\
  {\small\tt prakhar.gupta@epfl.ch } \\\And
  Matteo Pagliardini* \\
  Iprova SA, Switzerland \\
  {\small\tt mpagliardini@iprova.com}  \\\And
  Martin Jaggi \\
  EPFL, Switzerland \\
  {\small\tt martin.jaggi@epfl.ch} \\}

\date{}
\begin{document}
\maketitle

\begin{abstract}
  Pre-trained word vectors are ubiquitous in Natural Language Processing
  applications. In this paper, we show how training word embeddings jointly with
  bigram and even trigram embeddings, results in improved unigram embeddings.
  We claim that training word embeddings along with higher n-gram embeddings
  helps in the removal of the contextual information from the unigrams,
  resulting in better stand-alone word embeddings. We empirically show the
   validity of our hypothesis by outperforming other competing word
   representation models by a significant margin on a wide variety of tasks.
   We make our models publicly available.
\end{abstract}

\section{Introduction}
\blfootnote{* indicates equal contribution}
Distributed word representations are essential building blocks of modern NLP systems. Used as features in downstream applications, they often enhance generalization of models trained on a limited amount of data. They do so by capturing relevant distributional information about words from large volumes of unlabeled text.

Efficient methods to learn word vectors have been introduced in the past, most of them based on the distributional hypothesis of \citet{harris1954distributional,firth1957synopsis}: \textit{``a word is characterized by the company it keeps"}. While a standard approach relies on
global corpus statistics \cite{pennington2014glove} formulated as a matrix factorization using mean square reconstruction loss, other widely used methods are the bilinear word2vec architectures introduced by \citet{Mikolov2013EfficientEO}: While skip-gram aims to predict nearby words from a given word, CBOW predicts a target word from its set of context words.

Recently, significant improvements in the quality of the word embeddings were obtained by augmenting word-context pairs with sub-word information in the form of character n-grams \cite{bojanowski2017enriching}, especially for morphologically rich languages.
Nevertheless, to the best of our knowledge, no method has been introduced leveraging collocations of words with higher order word n-grams such as bigrams or trigrams as well as character n-grams together.

In this paper, we show how using higher order word n-grams along with unigrams during training can significantly
improve the quality of obtained word embeddings. The addition furthermore helps to disentangle contextual information present in the training data from the unigrams and results in overall better distributed word representations.

To validate our claim, we train two modifications of CBOW augmented with word-n-gram information during training. One is a recent sentence embedding method, Sent2Vec \cite{Pagliardini2018}, which we repurpose to obtain word vectors. The second method we propose is a modification of CBOW enriched with character n-gram information \cite{bojanowski2017enriching} that we again augment with word n-gram information. In both cases, we compare the resulting vectors with the most widely used word embedding methods on word similarity and analogy tasks and show significant quality improvements.
The code used to train the models presented in this
paper as well as the models themselves
 are made available to the public\footnote{publicly available on
\url{http://github.com/epfml/sent2vec}}.
\section{Model Description}

Before introducing our model, we recapitulate fundamental existing word embeddings methods.

\textbf{CBOW and skip-gram models}. Continuous bag-of-words (CBOW) and skip-gram models are standard log-bilinear models for obtaining word embeddings based on word-context pair information \cite{Mikolov2013EfficientEO}. Context here refers to a symmetric window centered on the target word $w_t$, containing the surrounding tokens at a distance less than some window size $\ws$: $C_t = \{w_k \,|\, k \in [t-\ws, t+\ws]\}$. The CBOW model tries to predict the target word given its context, maximizing the likelihood $\prod^T_{t=1}{p(w_t|C_t)}$, whereas skip-gram learns by predicting the context for a given target word maximizing $\prod^T_{t=1}{p(C_t|w_t)}$. To model those probabilities, a softmax activation is used on top of the inner product between a target vector $\uu_{w_t}$ and its context vector $\frac{1}{|C_t|}\sum_{w \in C_t}\vv_w$.

To overcome the computational bottleneck of the softmax for large vocabulary, negative sampling or noise contrastive estimation are well-established \cite{Mikolov:2013uz}, with the idea of employing simpler pairwise binary classifier loss functions to differentiate between the valid context $C_t$ and fake contexts $N_{C_t}$ sampled at random. While generating target-context pairs, both CBOW and skip-gram also use input word subsampling, discarding higher-frequency words with higher probability during training, in order to prevent the model from overfitting the most frequent tokens. Standard CBOW also uses a dynamic context window size: for each subsampled target word $w$, the size of its associated context window is sampled uniformly between 1 and $\ws$ \cite{Mikolov:2013uz}.

\textbf{Adding character n-grams}. \citet{bojanowski2017enriching} have augmented CBOW and skip-gram by adding character n-grams to the context representations. Word vectors are expressed as the sum of its unigram and average of its character n-gram embeddings $W_w$:

\[
\vv := \vv_w + \frac{1}{|W_w|}\sum_{c \in W_w}\vv_{c} \vspace{-2mm}
\]

Character n-grams are hashed to an index in the embedding matrix
. The training remains the same as for CBOW and skip-gram. This approach greatly improves the performances of CBOW and skip-gram on morpho-syntactic tasks. For the rest of the paper, we will refer to the CBOW and skip-gram methods enriched with subword-information as \emph{CBOW-char} and \emph{skip-gram-char} respectively.

\textbf{GloVe}. Instead of training online on local window contexts, GloVe vectors \cite{pennington2014glove} are trained using global co-occurrence statistics by factorizing the word-context co-occurrence matrix.

\textbf{Ngram2vec}. In order to leverage the performance of word vectors, training of word vectors using the skip-gram objective function with negative sampling is
augmented with n-gram co-occurrence information \cite{zhao2017ngram2vec}.

\subsection{Improving unigram embeddings by adding higher order word-n-grams to contexts}

\textbf{CBOW-char with word n-grams}. We propose to augment CBOW-char to additionally use word n-gram context vectors (in addition to char n-grams and the context word itself). More precisely, during training, the context vector for a given word $w_t$ is given by the average of all word-n-grams $N_t$, all char-n-grams, and all unigrams in the span of the current context window $C_t$:
\begin{equation}\label{eq:CharWordNgrams}
\vv := \frac{\sum_{w \in C_t}\!\vv_w +\! \sum_{n \in N_t}\!\vv_n +\! \sum_{w \in C_t}\sum_{c \in W_w}\!\vv_c}{|C_t|+|N_t|+\sum_{w \in C_t}|W_w|}
\end{equation}
For a given sentence, we apply input subsampling and a sliding context window as for standard CBOW. In addition, we keep the mapping from the subsampled sentence to the original sentence for the purpose of extracting word n-grams from the original sequence of words, within the span of the context window.
Word n-grams are added to the context using the hashing trick in the same way char-n-grams are handled. We use two different hashing index ranges to ensure there is no collision between char n-gram and word n-gram representations.

\textbf{Sent2Vec for word embeddings}. Initially implemented for sentence embeddings, Sent2Vec
 \cite{Pagliardini2018} can be seen as a derivative of word2vec's CBOW. The key differences between CBOW and Sent2Vec are the removal of the input subsampling, considering the entire sentence as context, as well as the addition of word-n-grams.

Here, word and n-grams embeddings from an entire sentence are averaged to form the corresponding sentence (context) embedding.

For both proposed CBOW-char and Sent2Vec models,
we employ dropout on word n-grams during training. For both models, word embeddings are obtained by simply discarding the higher order n-gram embeddings after training.

\begin{table*}[!htb]

\normalsize
\centering
\begin{tabular}{|l|c|c|c|}
\hline
         Model                            & WS 353                & WS 353 Relatedness    & \multicolumn{1}{l|}{WS 353 Similarity} \\ \hline
\textbf{CBOW-char}                & $.709 \pm .006$       & $.626 \pm .009$       & {\ul $.783 \pm .004$}                  \\
\textbf{CBOW-char + bi.}         & $.719 \pm .007$       & $.652 \pm .010$       & $.778 \pm .007$                        \\
\textbf{CBOW-char + bi. + tri.} & {\ul $.727 \pm .008$} & {\ul $.664 \pm .008$} & {\ul $.783 \pm .004$}                  \\ \hline
\textbf{Sent2Vec uni.}               & $.705 \pm .004$       & $.593 \pm .005$       & $.793 \pm .006$                        \\
\textbf{Sent2Vec uni. + bi.}         & $.755 \pm .005$       & $.683 \pm .008$       & $.817 \pm .007$                        \\
\textbf{Sent2Vec uni. + bi. + tri.}  & {\ul $\mathbf{.780 \pm .003}$} & {\ul $\mathbf{.721 \pm .006}$} & {\ul $\mathbf{.828 \pm .003}$}                  \\ \hline
\end{tabular}

  \vspace*{0.12 cm}

\begin{tabular}{|l|c|c|c|c|}
\hline
           Model                         & SimLex-999                     & MEN                            & \multicolumn{1}{l|}{Rare Words} & \multicolumn{1}{l|}{Mechanical Turk} \\ \hline
\textbf{CBOW-char}               & $.424 \pm .004$                & $.769 \pm .002$                & $.497 \pm .002$                 & $.675 \pm .007$                     \\
\textbf{CBOW-char + bi.}         & $.436 \pm .004$                & $.786 \pm .002$                & $.506 \pm .001$                 & $.671 \pm .007$                     \\
\textbf{CBOW-char + bi. + tri.}  & {\ul $.441 \pm .003$}          & {\ul $.788 \pm .002$}          & {\ul $\mathbf{.509 \pm .003}$}  & {\ul $.\mathbf{678 \pm .010}$}      \\ \hline
\textbf{Sent2Vec uni.}              & $.450 \pm .003$                & $.765 \pm .002$                & {\ul $.444 \pm .001$}           & $.625 \pm .005$                     \\
\textbf{Sent2Vec uni. + bi.}        & $.440 \pm .002$                & $.791 \pm .002$                & $.430 \pm .002$                 & {\ul $.661 \pm .005$}               \\
\textbf{Sent2Vec uni. + bi. + tri.} & {\ul $\mathbf{.464 \pm .003}$} & {\ul $\mathbf{.798 \pm .001}$} & $.432 \pm .003$                 & $.658 \pm .006$                     \\ \hline
\end{tabular}

  \vspace*{0.12 cm}

  \begin{tabular}{|l|c|c|c|}
\hline
             Model                        & \begin{tabular}[c]{@{}c@{}}Google\\ (Syntactic Analogies)\end{tabular} & \begin{tabular}[c]{@{}c@{}}Google\\ (Semantic Analogies)\end{tabular} & MSR                   \\ \hline
\textbf{CBOW-char}                & $.920 \pm .001$                                                        & {\ul $.799 \pm .004$}                                                   & { $.842 \pm .002$} \\
\textbf{CBOW-char + bi.}         & $.928 \pm .003$                                                        & $.798 \pm .006$                                                       & $.856 \pm .004$       \\
\textbf{CBOW-char + bi. + tri.} & {\ul $\mathbf{.929 \pm .001}$}                                                  & $.794 \pm .005$                                                       & {\ul $\mathbf{.857 \pm .002}$} \\ \hline
\textbf{Sent2Vec uni.}               & $.826 \pm .003$                                                        & $.847 \pm .003$                                                       & $.734 \pm .003$       \\
\textbf{Sent2Vec uni. + bi.}         & {\ul $.843 \pm .004$}                                                  & $.844 \pm .002$                                                       & {\ul $.754 \pm .004$} \\
\textbf{Sent2Vec uni. + bi. + tri.}  & $.837 \pm .003$                                                        & {\ul $\mathbf{.853 \pm .003}$}                                                 & { $.745 \pm .001$} \\ \hline
\end{tabular}

\caption{\textbf{Impact of using word n-grams}:
Models are compared using Spearman correlation measures for word similarity
 tasks and accuracy for word analogy tasks. Top performances on each dataset are
 shown in bold. An underline shows the best model(s) restricted to each architecture type. The abbreviations uni., bi., and tri. stand for unigrams, bigrams, and trigrams respectively.
 }\label{perf-boost}

\end{table*}

\section{Experimental Setup}

\subsection{Training}

We train all competing models on a wikipedia dump  of 69 million sentences containing 1.7 billion words, following \cite{Pagliardini2018}.

Sentences are tokenized using the Stanford NLP library \cite{manning2014stanford}.
All algorithms are implemented using a modified version of the fasttext
\cite{bojanowski2017enriching,Joulin2017BagOT} and sent2vec \cite{Pagliardini2018} libraries respectively.
Detailed training hyperparameters for all
models included in the comparison are provided in Table~\ref{table_train} in the
supplementary material. During training, we save models checkpoints at 20 equidistant intervals
and found out that the best performance for CBOW models occurs
around $60 - 80\%$ of the total training. As a result, we also indicate the checkpoint
at which we stop training the CBOW models. We use 300-dimension vectors for all our word embedding models.
For the Ngram2vec model, learning source and target embeddings for all the n-grams upto bigrams was
the best performing model and is included in the comparison.

For each method, we extensively tuned hyperparameters starting from the recommended values.
For each model, we select the parameters which give the best averaged results on our word-similarity and analogy tasks. After selecting the best hyperparameters, we train 5 models for each method, using a different random seed. The reported results are given as mean and standard deviation for those five models.
\begin{table*}[!htb]

\centering
\normalsize
\begin{tabular}{|l|c|c|c|}
\hline
Model                               & WS 353                & WS 353 Relatedness    & \multicolumn{1}{l|}{WS 353 Similarity} \\ \hline
\textbf{CBOW-char + bi. + tri.}     & $.727 \pm .008$       & $.664 \pm .008$       & $.783 \pm .004$                        \\
\textbf{Sent2Vec uni. + bi. + tri.} & $\mathbf{.780 \pm .003}$       & $\mathbf{.721 \pm .006}$       & $\mathbf{.828 \pm .003}$                        \\ \hline
\textbf{CBOW-char}                  & $.709 \pm .006$       & $.626 \pm .009$       & $.783 \pm .004$                        \\ \hline
\textbf{CBOW}                       & $.722 \pm .008$       & $.634 \pm .008$       & $.796 \pm .005$                        \\ \hline
\textbf{Skip-gram-char}             & $.724 \pm .007$       & $.655 \pm .008$       & $.789 \pm .004$                        \\ \hline
\textbf{Skip-gram}                  & $.736 \pm .004$       & $.672 \pm .007$       & $.796 \pm .005$                        \\ \hline
\textbf{GloVe}                      & $.559 \pm .002$ & $.484 \pm .005$ & $.665 \pm .008$                  \\ \hline
\textbf{Ngram2vec bi. - bi.}                      & $.745 \pm .003$ & $.687 \pm .003$ & $.797 \pm .004$                  \\ \hline
\end{tabular}

  \vspace*{0.12 cm}

  \begin{tabular}{|l|c|c|c|l|}
\hline
Model                               & SimLex-999            & MEN                   & Rare Words            & \multicolumn{1}{c|}{Mechanical Turk} \\ \hline
\textbf{CBOW-char + bi. + tri.}     & $.441 \pm .003$       & $.788 \pm .002$       & $\mathbf{.509 \pm .003}$       & $.678 \pm .010$                      \\
\textbf{Sent2Vec uni. + bi. + tri.} & $\mathbf{.464 \pm .003}$       & $\mathbf{.798 \pm .001}$       & $.432 \pm .003$       & $.658 \pm .006$                      \\ \hline
\textbf{CBOW-char}                  & $.424 \pm .004$       & $.769 \pm .002$       & $.497 \pm .002$       & $.675 \pm .007$                      \\ \hline
\textbf{CBOW}                       & $.432 \pm .004$       & $.757 \pm .002$       & $.454 \pm .002$       & $.674 \pm .006$                      \\ \hline
\textbf{Skip-gram-char}             & $.395 \pm .003$       & $.762 \pm .001$       & $.487 \pm .002$       & $\mathbf{.684 \pm .003}$                      \\ \hline
\textbf{Skip-gram}                  & $.405 \pm .001$       & $.770 \pm .001$       & $.468 \pm .002$       & $\mathbf{.684 \pm .005}$                      \\ \hline
\textbf{GloVe}                      & $.375  \pm .002$ & $.690 \pm .001 $ & $.327 \pm .002$ &    $.622 \pm .004$                                 \\ \hline
\textbf{Ngram2vec bi. - bi.}        & $.424  \pm .000$ & $.756 \pm .001 $ & $.462 \pm .002$ &    $.681 \pm .004$                                 \\ \hline
\end{tabular}

\vspace*{0.12 cm}

\begin{tabular}{|l|c|c|c|}
\hline
Model                               & \begin{tabular}[c]{@{}c@{}}Google\\ (Syntactic Analogies)\end{tabular} & \begin{tabular}[c]{@{}c@{}}Google\\ (Semantic Analogies)\end{tabular} & MSR                   \\ \hline
\textbf{CBOW-char + bi. + tri.}     & $\mathbf{.929 \pm .001}$                                                        & $.794 \pm .005$                                                       & $\mathbf{.857 \pm .002}$       \\
\textbf{Sent2Vec uni. + bi. + tri.} & $.837 \pm .003$                                                        & $\mathbf{.853 \pm .003}$                                                       & $.745 \pm .001$       \\ \hline
\textbf{CBOW-char}                  & $.920 \pm .001$                                                        & $.799 \pm .004$                                                       & $.842 \pm .002$       \\ \hline
\textbf{CBOW}                       & $.816 \pm .002$                                                        & $.805 \pm .005$                                                       & $.713 \pm .004$       \\ \hline
\textbf{Skip-gram-char}             & $.860 \pm .001$                                                        & $.828 \pm .005$                                                       & $.796 \pm .003$       \\ \hline
\textbf{Skip-gram}                  & $.829 \pm .002$                                                        & $.837 \pm .002$                                                       & $.753 \pm .005$       \\ \hline
\textbf{GloVe}                      & $.767 \pm .002$                                                  & $.697 \pm .007$                                            & $.678 \pm .003$  \\ \hline
\textbf{Ngram2vec bi. - bi.}                      & $.834 \pm .001$                                                  & $.812 \pm .003$                                            & $.761 \pm .001$  \\ \hline
\end{tabular}

\caption{\textbf{Improvement over existing methods}:
Models are compared using Spearman correlation measures for word similarity
 tasks and accuracy for word analogy tasks. Top performance(s) on each dataset is(are)
 shown in bold. The abbreviations uni., bi., and tri. stand for unigrams, bigrams, and trigrams respectively.
 }\label{overall-results}

\end{table*}

\subsection{Evaluation}

In order to evaluate our model, we use six datasets covering pair-wise word-similarity
tasks and two datasets covering word-analogy tasks.

\textbf{Word-similarity tasks}. Word-similarity tasks consist of word pairs along with
their human annotated similarity scores. To evaluate the performance of our models on
pair-wise word-similarity tasks, we use  \emph{WordSim353} (353 word-pairs)
\cite{finkelstein2002placing} divided into two datasets, \emph{WordSim Similarity} (203 word-pairs) and
\emph{WordSim Relatedness} (252 word-pairs) \cite{agirre2009study};  \emph{MEN} (3000 word-pairs)
\cite{bruni2012distributional}; \emph{Mechanical Turk} dataset \cite{radinsky2011word} (287 word-pairs);
\emph{Rare words dataset} (2034 word-pairs) \cite{luong2013better}; and \emph{SimLex-999} (999 word-pairs) \cite{hill2015simlex} dataset.

To calculate the similarity between two words, we use the cosine similarity between
their word representations. The similarity scores then, are compared to the human
ratings using Spearman's~$\rho$ \cite{Spearman1904proof} correlation scores.

\textbf{Word-analogy tasks}. Word analogy tasks pose analogy relations of the
form ``$x$ is to $y$ as $x^\star$ is to $y^\star$'', where $y$ is hidden and must be guessed
from the dataset vocabulary.

We use the \emph{MSR} \cite{mikolov2013linguistic} and the \emph{Google}
\cite{Mikolov2013EfficientEO} analogy datasets. The \emph{MSR} dataset contains
8000 syntactic analogy quadruplets while the \emph{Google} set has 8869 semantic
and 10675 syntactic relations.

To calculate the missing word in the relation, we use the 3CosMul
method \cite{levy2014linguistic}:\vspace*{-0.25 cm}
\begin{equation}\label{eq:3CosMul}
y^\star := \arg\!\max_{z \in \vocab \setminus\{x,y,x^\star\}} \frac{cos(\vv_{z},\vv_{y})\cdot cos(\vv_{z},\vv_{x^\star})}{cos(\vv_{z},\vv_{x}) + \varepsilon}
\end{equation}
\vspace*{0.1 cm}
where $\varepsilon = 0.0001$ is used to prevent division by 0 and $\vocab$ is the
dataset vocabulary.

We remove all the out of vocabulary words and are left with 6946 syntactic
relations for the \emph{MSR} dataset and 1959 word-pairs for the \emph{Rare Words} dataset. All other datasets do not have any out of vocabulary words.

\section{Results}

\textbf{Impact of word n-grams}. In Table \ref{perf-boost}, we evaluate the impact of adding contextual word n-grams to two CBOW variations: CBOW-char and Sent2Vec. By adding n-gram information, we consistently observe a boost in the Spearman correlation on the word similarity tasks. On the few datasets where we do not observe an improvement, the results for word-n-gram augmented methods are within standard deviation reach. The \emph{Rare Words} dataset for Sent2Vec is the only exception, despite getting some improvement for CBOW-char based methods. This observation can be attributed to the fact that character ngrams are shared between unigrams, enhancing generalization to infrequent words. Without char n-grams, the model might underfit those rare words, even more so with word n-grams.

We also see that the boost obtained by adding n-grams on word-similarity tasks is much larger for Sent2Vec models as compared to the CBOW-char ones possibly due to the fact that during training, Sent2Vec models use a much larger context and hence can use much more n-gram information for obtaining a better context representation.

For analogy tasks, we see an improvement in the augmented CBOW-char methods for morpho-syntactic analogy datasets with little or no gain for semantic analogy datasets. Yet, for Sent2Vec models, the gain is the other way around. This observation indicates the strong role played by character n-grams in boosting the performance on the syntactic tasks as well as restricting the word n-grams from improving the performance on semantic analogies.

\textbf{Comparison with competing methods}. In Table \ref{overall-results}, we compare word n-gram augmented methods with the most prominent word embedding models. We obtain state-of-the-art results for standalone unigram embeddings on most of the datasets confirming our hypothesis. The \emph{Mechanical Turk} dataset is the only exception.

We notice that Sent2Vec trigrams model dominates the word-similarity tasks as well as the semantic analogy tasks. However, character n-grams are quite helpful when it comes to syntactic analogy tasks underlining the importance of subword information. We also note that
the Ngram2vec model outperforms our augmented CBOW-char model in some of the tasks but is always inferior to Sent2Vec in those cases.

\section{Conclusion and Future Work}

We empirically show how augmenting the context representations using higher-order word n-grams improves the quality of word representations.
The empirical success also calls for a new theoretical model on the composite effect of training higher order n-grams simultaneously with unigrams.
Also, the success of Sent2Vec on word-level tasks, a method originally geared towards obtaining general purposed sentence embeddings,  hints
towards the additional benefits of using compositional methods for obtaining sentence/phrase representations.

\bibliography{naaclhlt2019}
\bibliographystyle{acl_natbib}

\appendix
\onecolumn
\newpage
\section{Training parameters for selected models}
\label{sec:param}
Training parameters for all  models except GloVe and Ngram2vec are provided in Table \ref{table_train}. For the GloVe model
, the minimum word count is set to 10; the window size is set to 10; we use 10 epochs for training; $X_{max}$, the weighting parameter for the word-context pairs is set to 100; all other parameters are set to default.
For Ngram2vec, the minimum word count is set to 10; the window size is set to 5; both source and target vectors  are trained for unigrams and bigrams; overlap
between the target word and source n-grams is allowed. All other features are set to default. To train the Ngram2vec models, we use the library provided by \cite{zhao2017ngram2vec}\footnote{\url{https://github.com/zhezhaoa/ngram2vec}}.
\begin{table*}[!htb]
\scriptsize

\centering
\begin{tabular}{|c|l|l|l|l|l|l|l|l|l|}
\hline
Model                                                                          & \multicolumn{1}{c|}{\begin{tabular}[c]{@{}c@{}}Sent2Vec\\ uni.\end{tabular}} & \multicolumn{1}{c|}{\begin{tabular}[c]{@{}c@{}}Sent2Vec\\ uni.+bi.\end{tabular}} & \multicolumn{1}{c|}{\begin{tabular}[c]{@{}c@{}}Sent2Vec\\ uni.+bi+tri.\end{tabular}} & \multicolumn{1}{c|}{\begin{tabular}[c]{@{}c@{}}CBOW\\ (char.)\end{tabular}} & \multicolumn{1}{c|}{\begin{tabular}[c]{@{}c@{}}CBOW\\(char.)+bi.\end{tabular}} & \multicolumn{1}{c|}{\begin{tabular}[c]{@{}c@{}}CBOW\\(char.)+bi.+tri.\end{tabular}} &
\multicolumn{1}{c|}{\begin{tabular}[c]{@{}c@{}}CBOW\end{tabular}} &
\multicolumn{1}{c|}{\begin{tabular}[c]{@{}c@{}}Skip-gram\\ (char.)\end{tabular}} &
\multicolumn{1}{c|}{\begin{tabular}[c]{@{}c@{}}Skip-gram\end{tabular}} \\ \hline
\begin{tabular}[c]{@{}c@{}}Embedding\\ Dimensions\end{tabular} & 300 & 300 & 300 & 300 & 300 & 300 & 300 & 300 & 300 \\ \hline
\begin{tabular}[c]{@{}c@{}}Max Vocab.\\ Size\end{tabular}  & 750k & 750k & 750k & 750k & 750k & 750k & 750k & 750k & 750k \\ \hline
\begin{tabular}[c]{@{}c@{}}Minimum\\ Word Count \end{tabular}  & 5 & 5 & 5 & 5 & 5 & 5 & 5 & 5 & 5 \\ \hline
\begin{tabular}[c]{@{}c@{}}Initial\\ Learning Rate\end{tabular}  & 0.2 & 0.2 & 0.2 & 0.05 & 0.05 & 0.05 & 0.05 & 0.05 & 0.05 \\ \hline
\begin{tabular}[c]{@{}c@{}}Epochs\end{tabular}  & 9 & 9 & 9 & 9 & 9 & 9 & 5 & 15 & 15 \\ \hline
\begin{tabular}[c]{@{}c@{}}Subsampling\\ hyper-param.\end{tabular}  & $1\times10^{-5}$ & $5\times10^{-5}$ & $5\times10^{-6}$ & $1\times10^{-4}$ & $1\times10^{-4}$ & $1\times10^{-4}$ & $1\times10^{-4}$ & $1\times10^{-4}$ & $1\times10^{-4}$ \\ \hline
\begin{tabular}[c]{@{}c@{}}Word-Ngrams\\ Bucket Size\end{tabular}  & - & 2M & 4M & - & 2M & 4M & - & - & - \\ \hline
\begin{tabular}[c]{@{}c@{}}Char-Ngrams\\ Bucket Size\end{tabular}  & - & - & - & 2M & 2M & 2M & - & 2M & - \\ \hline
\begin{tabular}[c]{@{}c@{}}Word-Ngrams\\ Dropped\\ per context\end{tabular}  & - & 4 & 4 & - & 2 & 2 & - & - & - \\ \hline
\begin{tabular}[c]{@{}c@{}}Window\\ Size\end{tabular}  & - & - & - & 10 & 10 & 10 & 10 & 5 & 5 \\ \hline
\begin{tabular}[c]{@{}c@{}}Number of\\ negatives\\ sampled\end{tabular}  & 10 & 10 & 10 & 5 & 5 & 5 & 5 & 5 & 5 \\ \hline
\begin{tabular}[c]{@{}c@{}}Max\\ Char-Ngram \\ Size\end{tabular}  & - & - & - & 6 & 6 & 6 & - & 6 & - \\ \hline
\begin{tabular}[c]{@{}c@{}}Min\\ Char-Ngram \\ Size\end{tabular}  & - & - & - & 3 & 3 & 3 & - & 3 & - \\ \hline
\begin{tabular}[c]{@{}c@{}}Percentage at\\ which training\\ is halted\\ (For CBOW\\ models only)\end{tabular}  & - & - & - & 75\% & 80\% & 80\% & 60\% & - & - \\ \hline
\end{tabular}
\caption{Training parameters for all non-GloVe models}\label{table_train}
\end{table*}

\end{document}